%% file: main.tex
\documentclass[10pt,twocolumn,letterpaper]{article}

\usepackage[pagenumbers]{cvpr}


\definecolor{cvprblue}{rgb}{0.21,0.49,0.74}
\usepackage[pagebackref,breaklinks,colorlinks,allcolors=cvprblue]{hyperref}

\input{math_commands.tex}

\usepackage{url}
\usepackage{multirow}
\usepackage{booktabs}
\usepackage{pifont}
\usepackage{xcolor}
\usepackage{graphicx}
\usepackage{amsmath}
\usepackage{algorithm}
\usepackage{mathtools}
\usepackage{algpseudocode}
\usepackage{listings}
\usepackage{wrapfig}
\usepackage{subcaption}
\usepackage{colortbl}
\usepackage{amssymb}
\usepackage{bbm}
\usepackage{gensymb}
\usepackage{xfrac}
\newcommand{\shline}{\noalign{\global\arrayrulewidth1pt}\hline\noalign{\global\arrayrulewidth0.4pt}}
\newcommand{\ord}{\mathrm{ord}}

\title{InstantSplat: Sparse-view SfM-free Gaussian Splatting in Seconds}

\author{
  Zhiwen Fan$^{1,2\dagger*}$,  
  Wenyan Cong$^{1*}$, 
  Kairun Wen$^{3*}$,
  Kevin Wang$^{1}$, 
  Jian Zhang$^{3}$, 
  Xinghao Ding$^{3}$, \\
  Danfei Xu$^{2,4}$, Boris Ivanovic$^{2}$, Marco Pavone$^{2,5}$, Georgios Pavlakos$^{1}$, Zhangyang Wang$^{1}$, Yue Wang$^{2,6}$ \\
  \small{*Authors contributed equally; $^\dagger$ Z. Fan is the Project Lead.} \\
  \normalsize{$^1$University of Texas at Austin \quad $^2$Nvidia Research \quad $^3$Xiamen University} \\
  \normalsize{$^4$Georgia Institute of Technology \quad $^5$Stanford University \quad $^6$University of Southern California} \\
  \normalsize{\textbf{\color{magenta} Project Website}: \url{https://instantsplat.github.io}} \\
}

\begin{document}
\twocolumn[{
\maketitle
\vspace{-5mm}
\begin{center}
    \includegraphics[width=0.98\textwidth]{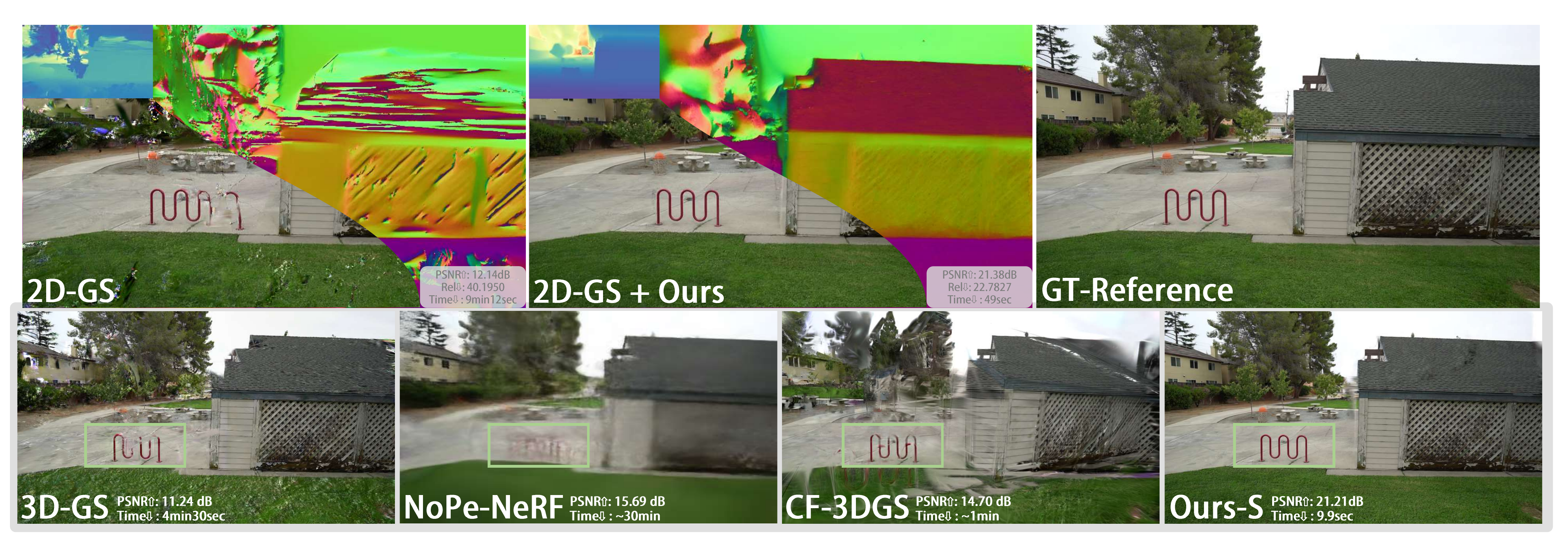}
    \vspace{-2mm}
    \captionof{figure}{\textbf{InstantSplat} processes sparse-view, unposed images to reconstruct a radiance field, capturing detailed scenes rapidly without relying on Structure-from-Motion. Optimization occurs under self-supervision with the support of the large pretrained model, MASt3R~\cite{leroy2024grounding}.}
    \label{fig:teaser}
\end{center}
\vspace{5mm}
}]

\begin{abstract}
\noindent
\textit{While neural 3D reconstruction has advanced substantially, it typically requires densely captured multi-view data with carefully initialized poses (e.g., using COLMAP). However, this requirement limits its broader applicability, as Structure-from-Motion (SfM) is often unreliable in sparse-view scenarios where feature matches are limited, resulting in cumulative errors. In this paper, we introduce \textbf{\textit{InstantSplat}}, a novel and lightning-fast neural reconstruction system that builds accurate 3D representations from as few as 2-3 images. InstantSplat adopts a self-supervised framework that bridges the gap between 2D images and 3D representations using Gaussian Bundle Adjustment (GauBA) and can be optimized in an end-to-end manner. InstantSplat integrates dense stereo priors and co-visibility relationships between frames to initialize pixel-aligned geometry by progressively expanding the scene avoiding redundancy. Gaussian Bundle Adjustment is used to adapt both the scene representation and camera parameters quickly by minimizing gradient-based photometric error. Overall, InstantSplat achieves large-scale 3D reconstruction in mere seconds by reducing the required number of input views. It achieves an acceleration of over 20 times in reconstruction, improves visual quality (SSIM) from 0.3755 to 0.7624 than COLMAP with 3D-GS, and is compatible with multiple 3D representations (3D-GS, 2D-GS, and Mip-Splatting).}
\end{abstract}

\input{sec/1_intro}

\input{sec/2_related}

\input{sec/3_method}
\input{sec/4_exp}
\input{sec/5_conclusion}

\input{sec/X_suppl}

\clearpage

{
    \small
    \bibliographystyle{ieeenat_fullname}
    \bibliography{main}
}

\end{document}

%% file: math_commands.tex
\usepackage{amsmath,amsfonts,bm}

\def\eqref#1{equation~\ref{#1}}

\def\1{\bm{1}}

\DeclareMathAlphabet{\mathsfit}{\encodingdefault}{\sfdefault}{m}{sl}
\SetMathAlphabet{\mathsfit}{bold}{\encodingdefault}{\sfdefault}{bx}{n}

\DeclareMathOperator*{\argmin}{arg\,min}

%% file: sec/1_intro.tex
\vspace{-4mm}
\section{Introduction}
\label{sec:intro}
3D reconstruction from a limited number of images has been a long-standing goal in computer vision. Conventional methods for 3D reconstruction utilize a complex and modular pipeline by decomposing this task into multiple subtasks, involving several stages of complex mappings between different data representations. For example, dense reconstruction using 3D Gaussian Splatting (3D-GS)~\cite{kerbl20233d} requires preprocessing from Structure-from-Motion (SfM), which first transforms images into key points, matches them, builds a scene graph, and incrementally reconstructs the scene representation while optimizing camera parameters and structures. Then, the subsequent 3D-GS assumes accurate camera parameters and performs a lengthy optimization process driven by a multi-view photometric loss, complemented by Adaptive Density Control (ADC), a heuristic that governs the creation or deletion of 3D primitives.
The processes of SfM and 3D-GS require densely captured multi-view images with large overlapping regions for feature matching~\cite{schonberger2016structure} and Gaussian point densification~\cite{kerbl20233d}, whereas the error derived from SfM can significantly degrade the 3D-GS optimization, poses challenges to the ADC process. As shown in Tab.~\ref{tab:sensitivity}, merely lowering the densification threshold or slightly perturb the initial camera parameters markedly affect the reconstruction quality. 

Significant efforts have been made to alleviate these strict requirements by either reducing the required image number or estimating camera poses in during optimization. Recent advancements in sparse-view reconstruction~\cite{jain2021putting,niemeyer2021regnerf,yang2023freenerf,wang2023sparsenerf,zhu2023fsgs} have shown notable progress by reducing the number of views from hundreds to just a few, though they still assume accurate camera poses obtained by leveraging dense views for pre-computation—an assumption that is rarely feasible.
Another line of research explores pose-free settings in NeRFmm~\cite{wang2021nerf}, Nope-NeRF~\cite{bian2023nope}, and CF-3DGS~\cite{fu2023colmap}. These approaches also assume dense data coverage, often sourced from video sequences. Such dense data requires an extensive optimization process not only assumes the known camera intrinsics~\cite{bian2023nope,fu2023colmap}, but also usually taking \underline{hours} to reconstruct a single 3D scene.

In this paper, we present \textbf{InstantSplat}, a system that builds large-scale 3D representations from sparse-view data, offering two distinct advantages: (1) it generalizes well to various scene types with minimal multi-view coverage (as few as 2 or 3 images) for high-quality reconstruction, and (2) it is lightning-fast thanks to its self-supervised paradigm with gradient-based photometric error. 

To achieve this, InstantSplat switches from previous system incorporating sparse SfM with complex ADC~\cite{kerbl20233d} to a end-to-end optimization framework aided by geometric priors. Specifically, it introduces Gaussian Bundle Adjustment to link 2D and 3D, rendering 2D images from 3D representations and computing gradients based on the residual between rendered images and ground-truth images, jointly optimizing both the scene representation and camera parameters.
To facilitate convergence, InstantSplat incorporates generalizable geometry priors from MASt3R~\cite{leroy2024grounding} to obtain densely covered, pixel-aligned stereo points and subsequently adopts a co-visibility cross-view check to initialize the scene representation by expanding the points. The end-to-end framework requires only a few optimization steps with an adaptive confidence-aware learning rate, constructing an explicit 3D representation in just seconds.
We evaluate InstantSplat on the Tanks and Temples~\cite{knapitsch2017tanks} and MVImgNet~\cite{yu2023mvimgnet} datasets. InstantSplat achieves a dramatic reduction in optimization time to just 7.5 seconds while significantly improving SSIM from 0.3755 to 0.7624 in a 3-view setting, and it further enhances pose accuracy. Moreover, InstantSplat is compatible with other neural 3D representations, such as 2D-GS~\cite{Huang2DGS2024} for surface reconstruction and Mip-Splatting~\cite{yu2024mip} for multi-resolution reconstruction, highlighting its robust generalization capability.

In summary, our main contributions are:
\begin{itemize} 
\item We propose a novel system for neural 3D reconstruction that transitions from traditional SfM, which rely on lengthy 3D-GS optimizations, to a rapid and self-supervised framework. It leverages geometric priors and joint optimization to significantly reduce the required number of views, while effectively generalizing to diverse point-based representations.
\item We introduce Gaussian Bundle Adjustment to connect 2D and 3D, establishing an end-to-end paradigm to adjust scene representation and camera parameters based on gradient-based photometric error. 
\item We propose initializing the framework with a learnable dense model, equipped with co-visibility expansion to initialize dense surface points. 
\end{itemize}

%% file: sec/2_related.tex
\section{Related Works}

\begin{table}[t]
\centering
\resizebox{0.88\columnwidth}{!}{
\begin{tabular}{@{}l|cc}
  & SSIM $\uparrow$ & LPIPS~$\downarrow$ \\
\hline
COLMAP+3DGS~\cite{kerbl20233d} & 0.7633 & 0.2122 \\ \hline
\quad + Adjust ADC Threhold.  & 0.7817$_\text{(\textcolor{blue}{\textbf{+0.0184}})}$ & 0.1666$_\text{(\textcolor{blue}{\textbf{-0.0456}})}$\\
\quad + Mask 30\% SfM Points & 0.7406$_\text{(\textcolor{red}{\textbf{-0.0227}})}$ & 0.2427$_\text{(\textcolor{red}{\textbf{+0.0305}})}$\\
\quad $\pm$ 1$^\circ$ Noise on Rotation  & 0.3529$_\text{(\textcolor{red}{\textbf{-0.4104}})}$ & 0.5265$_\text{(\textcolor{red}{\textbf{+0.3143}})}$ \\
\end{tabular}%
}
\caption{\textbf{Sensitivity Analysis in Adaptive Density Control (ADC)}. The 3D-GS~\cite{kerbl20233d} utilizes camera parameters and sparse points from Structure-from-Motion (SfM). Adjusting the densification gradient threshold, an ADC parameter~\cite{kerbl20233d}, to half significantly enhances rendering quality (LPIPS: 0.2122 $\mapsto$ 0.1666). Conversely, non-uniform point distribution through 30\% downsampling poses challenges (LPIPS: 0.2122 $\mapsto$ 0.2427), and slight rotational perturbations in SfM poses lead to substantial performance drops (LPIPS: 0.2122 $\mapsto$ 0.5265). Experiments conducted on the ``Bicycle'' scene from the MipNeRF360~\cite{Barron2022MipNeRF3U} dataset.}
\vspace{-2mm}
\label{tab:sensitivity}
\end{table}

\paragraph{\textbf{3D Representations}}
Novel view synthesis aims to generate unseen views of an object or scene from a given set of images~\cite{avidan1997novel,mildenhall2019local}. Neural Radiance Fields (NeRF)\cite{mildenhall2021nerf}, enables photo-realistic rendering quality, employs MLPs to represent 3D scenes, taking the directions and positions of 3D points as input and employing volume rendering for image synthesis. Despite its popularity, NeRF faces challenges in terms of speed during both training and inference phases. Subsequent enhancements primarily focus on either enhancing quality\cite{Barron2022MipNeRF3U,barron2022mip,barron2023zip} or improving efficiency~\cite{penner2017soft,seitz2002photorealistic,srinivasan2020lighthouse,srinivasan2019pushing}, with few achieving both.
Recent advancements in unstructured radiance fields\cite{chen2023neurbf,kerbl20233d,xu2022point} introduce a set of primitives to represent scenes. Notably, 3D Gaussian Splatting (3D-GS)~\cite{kerbl20233d} uses anisotropic 3D Gaussians~\cite{zwicker2001ewa} to depict radiance fields, coupled with differentiable splatting for rendering. This method has shown considerable success in rapidly reconstructing complex real-world scenes with high quality. And many works extend 3D-GS for surface reconstruction~\cite{huang20242d,yu2024gaussian,zhang2024rade}, neural gaussians~\cite{lu2024scaffold}, multi-resolution modeling~\cite{yu2024mip,feng2024flashgs} and feed-forward reconstruction~\cite{charatan2024pixelsplat,chen2025mvsplat,fan2024large,xu2024depthsplat,hong2024pf3plat,li2024ggrt,ye2024no,ziwen2024long,jin2024lvsm}. However, these point-based representations involves a large number of hyperparameters for different datasets, especially for the adaptive density control, which servers as the core function to densify from sparse SfM point cloud to densely covered 3D Gaussian points. However, the optimization of SfM is independent with the followed 3D Gaussian optimization, and the accumulated error cannot be mitigated with current pipelines. This motivate us to design a holistic, efficient and robust framework to represent the 3D world.
\vspace{-2mm}

\begin{figure*}[t!]
\vspace{-1em}
\vskip 0.2in
\begin{center}
\centering
\includegraphics[width=0.99\linewidth]{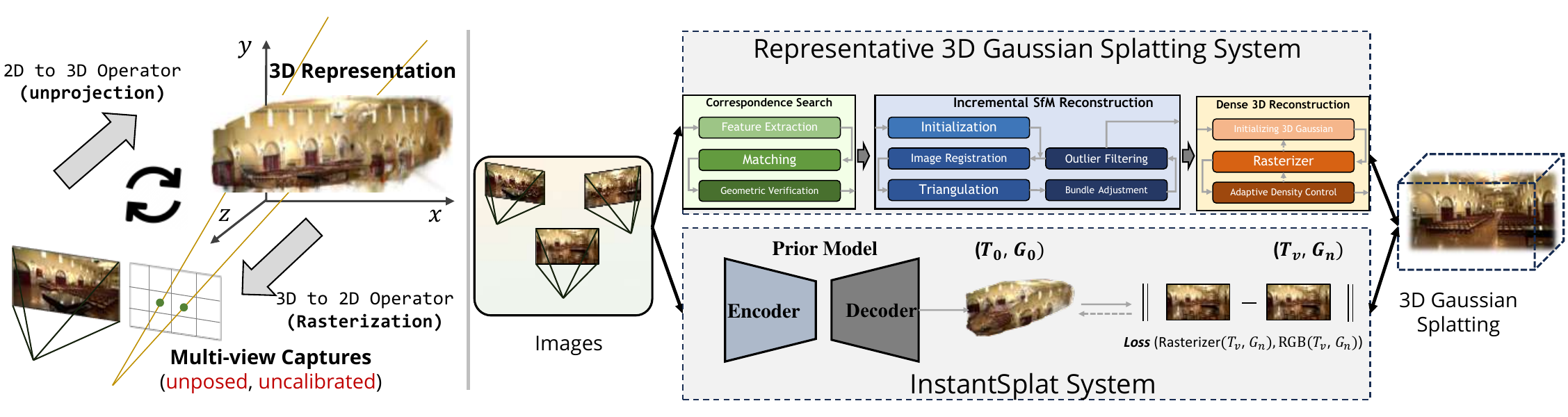}
\vspace{-3mm}
\caption{\textbf{Overall Framework of InstantSplat.} Unlike the modular COLMAP pipeline with Gaussian Splatting, which relies on time-consuming and accuracy-sensitive ADC processes within 3D-GS, and accurate camera poses and sparse point clouds from SfM, InstantSplat employs a deep model to initialize dense surface points. It adopts an end-to-end framework that iteratively optimizes both 3D representation and camera poses, enhancing efficiency and robustness.}\label{fig:arc}
\vspace{-5mm}
\end{center}
\end{figure*}

\vspace{-3mm}
\paragraph{\textbf{Unconstraint Novel View Synthesis}}
NeRFs and 3DGS require carefully captured hundreds of images to ensure sufficient scene coverage as input and utilize preprocessing Structure-from-Motion (SfM) software, such as COLMAP~\cite{schonberger2016structure}, to compute camera parameters, and sparse SfM point cloud as additional inputs. 
However, the densely captured images with the strong reliance on COLMAP significantly limits practical applications, it requires the users with expertise in the field of photography and requires significant computing resources (hours for each individual scene). The accumulated error from SfM will propogate to the following differential 3D representation and SfM may fail with captured images without sufficient overlappings and rich textures.  
To address the challenge of the requisite number of views, various studies have introduced different regularization techniques to optimize the radiance fields. For instance, Depth-NeRF~\cite{deng2021depth} employs additional depth supervision to enhance rendering quality. RegNeRF~\cite{niemeyer2021regnerf} and SparseNeRF~\cite{wang2023sparsenerf} introduces a depth prior loss for geometric regularization. DietNeRF~\cite{jain2021putting}, SinNeRF~\cite{xu2022sinnerf}, and ReconFusion~\cite{wu2023reconfusion} leverages supervision in the CLIP/DINO-ViT/Diffusion Model to constrain the rendering of unseen views. 
PixelNeRF~\cite{yu2021pixelnerf} and FreeNeRF~\cite{yang2023freenerf} utilize pre-training and frequency annealing for few-shot NeRF.
FSGS~\cite{zhu2023fsgs}, and SparseGS~\cite{xiong2023sparsegs} employ monocular depth estimators or diffusion models on Gaussian Splatting for sparse-view conditions.
However, these methods require known \underline{ground-truth camera poses} computed and sampled from using dense views in the preprocessing, and Structure-from-Motion (SfM) algorithms often fail to predict camera poses and point cloud with sparse inputs due to insufficient image correspondences.
Therefore, another line of research focuses on pose-free 3D optimization with uncalibrated images as direct input. NeRFmm~\cite{wang2021nerf} simultaneously optimizes camera intrinsics, extrinsics, and NeRF training. BARF~\cite{lin2021barf} introduces a coarse-to-fine strategy for encoding camera poses and joint NeRF optimization. SCNeRF~\cite{jeong2021self} adds camera distortion parameterization and employs geometric loss for ray regularization. Similarly, GARF~\cite{jeong2021self} demonstrates that Gaussian-MLPs facilitate more straightforward and accurate joint optimization of pose and scene. SPARF~\cite{truong2023sparf} adds Gaussian noise to simulate noisy camera poses. Recent works, such as Nope-NeRF~\cite{bian2023nope}, Lu-NeRF~\cite{cheng2023lu}, LocalRF~\cite{meuleman2023progressively} and CF-3DGS~\cite{fu2023colmap}, leverage depth information to constrain NeRF or 3DGS optimization. The work in~\cite{jiang2024construct} utilizes monocular depth estimator and additional image matching network try to reduce the view number, with roughly hours of optimization for a single scene.
These pose-free works generally presume the input are \underline{dense video sequences}~\cite{fu2023colmap,bian2023nope} with known viewing order and camera intrinsics~\cite{jiang2024construct,fu2023colmap,bian2023nope}, and the optimization for each scene (usually several hours) is even longer than COLMAP with NeRF or 3D-GS variants.

In contrast, InstantSplat proposes reconstructing a 3D neural representation using only 2D images, jointly optimizing camera poses and the 3D model aided by geometric priors from the large-scale trained model, MASt3R~\cite{leroy2024grounding}. We categorize InstantSplat as a self-supervised framework following INeRF's definition~\cite{yen2021inerf}, where it employs a trained NeRF model to estimate camera poses.

%% file: sec/3_method.tex
\section{Method}
\paragraph{\textbf{Overview}}
Our optimization framework is a self-supervised structure that leverages photometric error to guide the training of the 3D representation and camera poses, as illustrated in Fig.~\ref{fig:arc}. In Sec.~\ref{sec:prelim}, we provide background on 3D-GS and the 3D prior model. Following this, in Sec.~\ref{sec:self-sup}, we describe our self-supervised training scheme and the development of the end-to-end optimizable framework. For the initialization of the 3D representation, we utilize a prior model to obtain regressed point maps and camera parameters, enhanced by a co-visibility-based geometric initialization (Sec.~\ref{sec:co-visible}). The overall gradient-based joint optimization framework utilizes photometric error to align 3D Gaussians with 2D images, employing a confidence-aware optimizer with few optimization steps (Sec.~\ref{sec:optimization}).

\subsection{Preliminary}~\label{sec:prelim}
\textbf{3D Gaussian Splatting (3D-GS)}~\cite{kerbl20233d} is an explicit 3D scene representation utilizing a set of 3D Gaussians to model the scene. A 3D Gaussian is parameterized by a mean vector $\mathbf x \in \mathbb{R}^3$, an opacity $\alpha$ and a covariance matrix $\Sigma\in \mathbb{R}^{3\times 3}$:
\begin{equation}
    G(\mathbf p, \alpha, \Sigma)=\alpha \exp \left(-\frac{1}{2}{\left(\mathbf p-\mathbf x\right)}^T\Sigma^{-1}\left(\mathbf p-\mathbf \mu\right)\right)\label{eq:3dgs}
\end{equation}
To represent the view-direction-dependent, spherical harmonic (SH) coefficients are attached to each Gaussian, and the color is computed via $\mathbf c\left(\mathbf d\right) = \sum_{i=1}^n {\mathbf c}_i \mathcal{B}_i\left(\mathbf d\right)$, where $\mathcal{B}_i$ is the $\ord_i$ SH basis. And the color is rendered via $\mathbf c = \sum_{i=1}^n\mathbf c_i \alpha_i \prod_{j=1}^{i-1}(1-\alpha_j)$, where $\mathbf c_i$ is the color computed from the SH coefficients of the $\ord i$ Gaussian. $\alpha_i$ is given by evaluating a 2D Gaussian with covariance multiplied by a learned per-Gaussian opacity. 
The 2D covariance matrix is calculated by projecting the 3D covariance to the camera coordinate system. The 3D covariance matrix is decomposed into a scaling matrix and a rotation matrix.
In summary, 3D-GS uses a set of 3D Gaussians $\left\{\mathbf{G}_i|i=1,2,...,n\right\}$ to represent a scene, where each 3D Gaussian $\mathbf{G}_i$ is characterized by: position $\mathbf x\in  \mathbb{R}^3$, a series of SH coefficients $\left\{\mathbf c_i\in \mathbb{R}^{3}|i=1,2,...,n\right\}$, opacity $\alpha\in  \mathbb{R}$, rotation $\mathbf q \in  \mathbb{R}^4$ and scaling $\mathbf s \in  \mathbb{R}^3$.

Distinguished from 3D-GS, 2D-GS~\cite{Huang2DGS2024} proposes that reducing one scaling dimension to zero and approximating the primitive as a surface can improve view consistency. To implement this, 2D-GS replaces the volumetric representations used in 3D-GS with planar disks, which are defined within the local tangent plane and mathematically expressed as follows:
\begin{equation}
    G(\mathbf p)=\exp\left(-\frac{u(\mathbf p)^2+v(\mathbf p)^2}2\right) \label{eq:2dgs}
\end{equation}
where $u(\mathbf p)$ and $v(\mathbf p)$ are local $UV$ space coordinates.

During the rendering process, 2D-GS substitutes direct affine transformations with three non-parallel planes to define ray-splat intersections. Following this, rasterization is applied after a low-pass filter to produce smoother results.

\textbf{MASt3R} is a feed-forward Multi-View Stereo model where it regresses the pixel-aligned point maps directly from raw images. It utilize a Transformer architecture with cross-view information flow to predict the $\hat{\mathbf{P}}_{1,1}$ and $\hat{\mathbf{P}}_{2,1}$, obtained from two corresponding views $\{ 1, 2\}$ where the camera origin as view 1, on which the ground-truth is defined. 
The regression loss for training MASt3R is as: 
\begin{equation}
\mathcal{L}_{\text{reg}} =  \left\Vert \frac{1}{z_i} \cdot \mathbf{P}_{v,1}  -   \frac{1}{z_i} \cdot \hat{\mathbf{P}}_{v,1} \right\Vert\label{eq:regression}
\end{equation}
The view $v \in \{1, 2 \}$, $\mathbf{P}$ and $\hat{\mathbf{P}}$ are the predicted and ground-truth point maps, separately. 
$z_i = \text{norm}(\mathbf{P}_{1,1}, \mathbf{P}_{2,1})$ is the normalization factor.
It also regresses the pixel-aligned confidence score $ \mathbf{O}_{v,1}^i $ and optimizes as:
\begin{equation}
 \mathcal{L}_{\text{conf}} = \sum_{v \in \{1,2\}} \, \sum_{i \in \mathbf{D}^v} 
            \mathbf{O}_{v,1}^i \cdot \mathcal{L}_{\text{reg}}(v,i) - \alpha \cdot  \log \mathbf{O}_{v,1}^i,
\label{eq:conf_loss}
\vspace{-2mm}
\end{equation}
where $\alpha$ is a hyper-parameter controlling the regularization term, encouraging the network to extrapolate in harder areas.

\subsection{Optimizating 3D with Self-supervision}~\label{sec:self-sup}

\vspace{-5mm}
\paragraph{Gaussian Bundle Adjustment}
Given multiple views and a 3D representation with initialized poses, represented by a set of Gaussians \(\mathbf{G}\), and \(\mathbf{T}\), 
we employ the pre-built rasterizer from Gaussian Splatting as a differentiable operator to render the images corresponding to the target poses \(\mathbf{T}\).  
Gradient descent with photometric error minimization is used to reduce the discrepancy between the optimizable 3D representation and 2D observations, iteratively refining \(\mathbf{G}\), and \(\mathbf{T}\) toward representative orientations:
\vspace{-2mm}
\begin{align} \label{eqn:photo_loss}
\footnotesize
\mathbf{G}^{*},\mathbf{T}^{*} = \argmin_{\mathbf{G},\mathbf{T}} \sum_{v \in N}  \sum_{i=1}^{HW} & \left\Vert \tilde{\mathbf{C}}_{v}^{i}(\mathbf{G},\mathbf{T}) -\mathbf{C}_{v}^{i}(\mathbf{G},\mathbf{T}) \right\Vert   \nonumber
\end{align}
The $\mathbf{C}$ denotes the rasterization function applied to $\mathbf{G}$ at $\mathbf{T}$, while $\tilde{\mathbf{C}}$ represents the observed 2D images.

We initialize $\mathbf{G}$ and $\mathbf{T}$ using a post-processing step following the pairwise predictions inferred by MASt3R, which we will elaborate on later. 

\vspace{-3mm}
\paragraph{\textbf{Aligning Camera Poses on Test Views}}
Unlike benchmark datasets where camera pose estimation and calibration are pre-computed with access to all training and testing views, a pose-free setting requires additional steps to regress the poses for the test views. Similar to INeRF~\cite{wang2021nerf}, we freeze the 3D model learned by training images, while optimizing the camera poses for the test views. This optimization process focuses on minimizing the photometric discrepancies between the synthesized images and the actual images on test views, aiming to achieve alignment between the 3D model and test images for evaluating visual quality and pose estimation metrics.

\subsection{Co-visible Global Geometry Initialization}\label{sec:co-visible}

\paragraph{Scaling Pairwise to Global Alignment}
With the off-the-shell stereo model, MASt3R, we regress a set of stereo point maps along with calculated camera intrinsics and extrinsics following Sec.~\ref{sec:prelim}. To align these pair wise predictions into a globally aligned results, we follow MASt3R to build a complete connectivity graph $\mathcal{G}(\mathcal{V},\mathcal{E})$ of all the N input views, where the vertices $\mathcal{V}$ and each edge $e=(n,m) \in \mathcal{E}$ indicate a shared visual content between images $I_n$ and $I_m$. To convert the initially predicted point map $\{ (\mathbf{P}_i \in \mathbb{R}^{H \times W \times 3}) \}_{i=1}^{N}$ to be a globally aligned one $\{ (\mathbf{\tilde{P}}_i \in \mathbb{R}^{H \times W \times 3}) \}_{i=1}^{N}$, we follow the MASt3R to update these point maps, transformation matrix, and a scale factor into globally corrdinate system. Please refer to the supplementary material for detailed description.
However, the pairwise predicted focal lengths remain inconsistent post-optimization. Assuming a single-camera setup akin to COLMAP for capturing a scene, we stabilize the estimated focal lengths by averaging them across all training views: $\bar{f} = \frac{1}{N} \sum_{i=1}^{N} f_i^*$ which is simple but significantly improve optimized visual quality.

\vspace{-3mm}
\paragraph{Co-visible Redundancy Elimination}
Utilizing point maps from all training images typically results in redundancy, manifested through overlapping regions that produce duplicated, pixel-aligned point clouds. This not only increases the free parameters in scene representation but also slows down the optimization process. To combat this, we exploit co-visibility across training images to eliminate these duplicate pixels. \textbf{View ranking} is conducted by computing the average confidence levels from the confidence maps of each image, with a higher average indicating a more accurate initial geometry from the 3D prior model. We prune points from views with lower confidence scores to enhance point quality, measured by the view confidence score,
\begin{equation}
     v_i = \frac{1}{|\mathbf{O}_i|} \sum_{o \in \mathbf{O}_i} o
\end{equation}
where $v_i$ denotes the average confidence score of view $i$.

\textbf{Cross-view checking} starts with the view ($j$) having the lowest average confidence, where we combine point maps from views with higher average confidence score than view $j$, $\{ (\mathbf{\tilde{P}}_i \in \mathbb{R}^{H \times W \times 3}) \}_{i=1, i \neq j, v_i > v_j}^{N}$, and project them onto view $j$. Depth ambiguity is addressed through a depth verification process, pruning points only if the difference between the projected and original depth values is higher than a predefined threshold: \( \Delta d = |d_{\text{proj}} - d_{\text{orig}}| \). This procedure is iterated in ascending order across all images, ultimately generating a binary mask $\{ (\mathbf{\tilde{M}}_i \in \mathbb{R}^{H \times W \times 1}) \}_{i=1}^{N}$ to filter out redundant points.
More formally, the cross-view co-visibility check for removing redundancy in view $j$ can be written as:
\begin{align}
\mathbf{D}_{\text{proj},j} &= \bigcap_{\{i \mid v_i > v_j\}} \text{Proj}_j(\mathbf{\tilde{P}}_i) \\
\mathcal{M}_j &= \begin{cases} 
1 & \text{if } |\mathbf{D}_{\text{proj},j} - \mathbf{D}_{\text{orig},j}| < \theta \\
0 & \text{otherwise}
\end{cases} \\
\mathbf{P}_j &= (1 - \mathcal{M}_j) \cdot \mathbf{\tilde{P}}_j
\end{align}
where \text{Proj} is the function applying the projection operator across all views except view $i$ ($v_i \geq v_j$).
$\mathcal{M}_j$ is the visibility mask, determined by comparing the projected depth $d_{\text{proj},j}$ to the original depth $d_{\text{orig},j}$. Points with depth differences below a threshold $\theta$ are considered visible and consistent, thus redundant.
We use inverse of $\mathcal{M}_j$ to mask out redundant points.

\subsection{Optimization}~\label{sec:optimization}

\vspace{-10mm}
\paragraph{Confidence-aware Optimizer}
In our pursuit to refine the accuracy of scene reconstruction, we introduce a confidence-aware per-point optimizer designed to facilitate convergence during self-supervised optimization. This approach is tailored to dynamically adjust learning rates based on the per-point confidence as detailed in Sec.~\ref{sec:co-visible}. By prioritizing points with lower confidence, which are more susceptible to errors, we ensure a targeted and efficient optimization process. The calibration of confidence scores involves normalizing them into a proper range, thereby addressing points in need of significant corrections. The normalization and adjustment process is mathematically outlined as follows:
$\mathbf{O}_{\text{norm}} = (1 - \text{sigmoid}(\mathbf{O}_{\text{init}})) \cdot \beta $
where $\beta$ is a hyperparameter used to fine-tune the scaling effect, optimizing the learning rates to enhance the convergence.

\vspace{-3mm}
\paragraph{Training Objective}
To optimize our neural 3D representation efficiently, we employ purely visual signals, using photometric loss to minimize discrepancies between the rasterized images from our 3D model, denoted as $\tilde{\mathbf{C}}(\mathbf{G}, \mathbf{T})$, and the observed images $\mathbf{C}$. 
Overall, the process maps observed unposed RGB images, to a 3D representation $\mathbf{G^*}$ with accurate pose estimations $ \mathbf{T^*}$, utilizing end-to-end iterative optimization and a self-supervised training objective.
InstantSplat aligns scene representations and camera poses toward visual observation, improving rendering accuracy and optimization efficiency without additional heuristic regularization.

\begin{figure*}[t!]
\vspace{-1em}
\vskip 0.2in
\begin{center}
\centering
\includegraphics[width=0.99\linewidth]{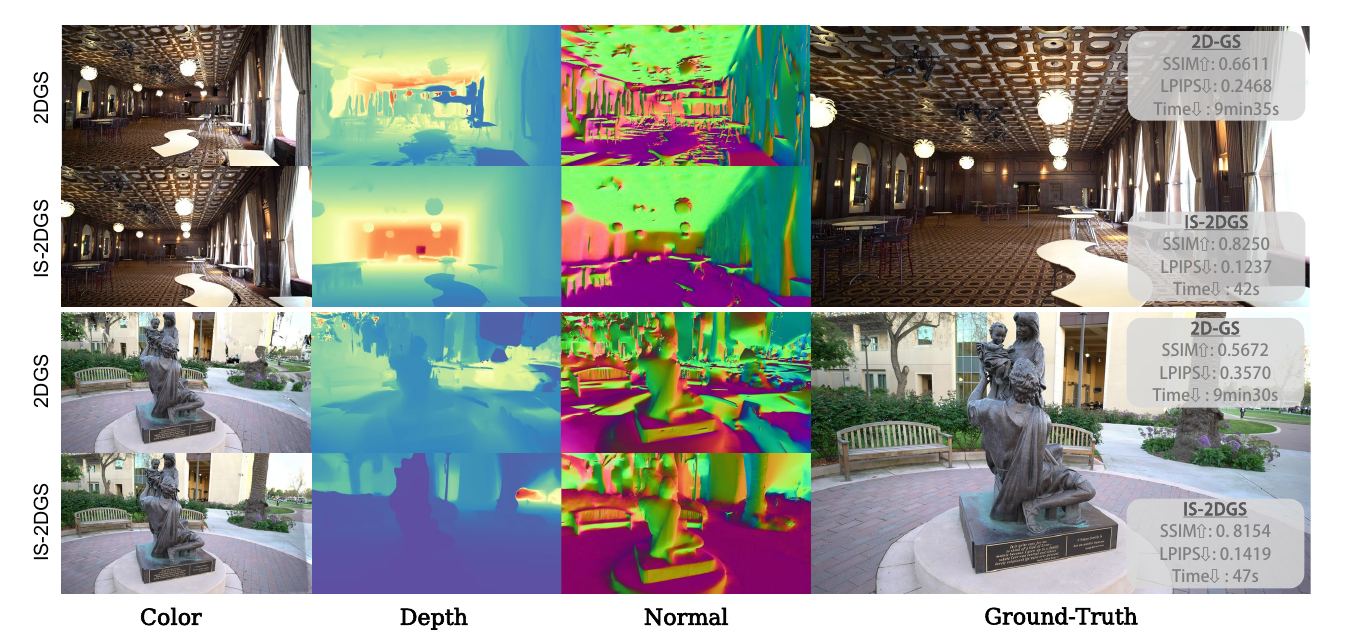}
\caption{\textbf{Visual Comparisons between 2D-GS and InstantSplat-2D-GS under 3-view setting}. We present the rendering results of RGB images, depth maps, and normal maps from the Tanks and Temples dataset. The corresponding SSIM, LPIPS, and training time metrics are displayed on
the right. Here, IS is used as an abbreviation for InstantSplat.}
\label{fig:main_exp_geo}
\end{center}
\end{figure*}
mechanism efficiently adapts parameters through the use of photometric loss, applicable to both camera and Gaussian parameters.
\begin{table*}
\centering
\caption{\textbf{Quantitative Evaluations on Tanks and Temples Datasets.} Our method achieves significantly clearer details (lower LPIPS) compared to other pose-free methods, even with an optimization time of only $\sim$7.5 seconds (\textbf{Ours-S}), and avoids artifacts caused by noisy pose estimation, as quantified by Absolute Trajectory Error (ATE). Extending training time further improves rendering quality. InstantSplat also demonstrates superior camera pose estimation accuracy in ATE with ground truth scale. In 3D-GS, ADC refers to Adaptive Density Control~\cite{kerbl20233d}. We further leverage \href{https://pypi.org/project/cupy/}{\textbf{cupy}} to accelerate MASt3R inference.}
\label{tab:tt_eval}
\resizebox{1.99\columnwidth}{!}{
\begin{tabular}{l|ccc|ccc|ccc|ccc}
 &\multicolumn{3}{c|}{Training Time}   & \multicolumn{3}{c|}{SSIM}   & \multicolumn{3}{c|}{LPIPS}  & \multicolumn{3}{c}{ATE}            \\
   & 3-view   & 6-view  & 12-view    & 3-view   & 6-view  & 12-view   & 3-view   & 6-view  & 12-view     & 3-view   & 6-view  & 12-view             
\\ \shline
COLMAP + 3DGS~\cite{kerbl20233d}                  
& 4min28s & 6min44s & 8min11s & 0.3755 & 0.5917 & 0.7163 & 0.5130 & 0.3433 & 0.2505 & - & - & -
\\
COLMAP + FSGS~\cite{zhu2023fsgs}
 & 2min37s & 3min16s & 3min49s & 0.5701 & 0.7752 & 0.8479 & 0.3465 & 0.1927 & 0.1477 & - & - & -
\\ \hline

NoPe-NeRF~\cite{bian2023nope}  
& 33min & 47min & 84min & 0.4570 & 0.5067 & 0.6096  & 0.6168  & 0.5780 & 0.5067 & 0.2828 & 0.1431 & 0.1029
\\
CF-3DGS~\cite{fu2023colmap}
  & 1min6s & 2min14s & 3min30s & 0.4066 & 0.4690 & 0.5077 & 0.4520 & 0.4219 & 0.4189 & 0.1937 & 0.1572 & 0.1031
\\\hline
NeRF-mm~\cite{wang2021nerf}                    
& 7min42s & 14min40s & 29min42s & 0.4019 & 0.4308 & 0.4677 & 0.6421 & 0.6252 & 0.6020 & 0.2721 & 0.2329 & 0.1529
\\\hline
\textbf{Our-S}                                                     
& \textbf{7.5s} & \textbf{13.0s} & \textbf{32.6s} &  \textbf{0.7624} & 0.8300 & 0.8413 & 0.1844 & 0.1579 & 0.1654 & 0.0191 & 0.0172 & 0.0110
\\
\textbf{Our-XL}              
& 25.4s & 29.0s & 44.3s & 0.7615 & \textbf{0.8453} & \textbf{0.8785} & \textbf{0.1634} & \textbf{0.1173} & \textbf{0.1068} & \textbf{0.0189} & \textbf{0.0164} & \textbf{0.0101}
\\ 
\end{tabular}}
\end{table*}

\begin{figure*}[t!]
\vspace{-1em}
\vskip 0.2in
\begin{center}
\centering
\includegraphics[width=0.99\linewidth]{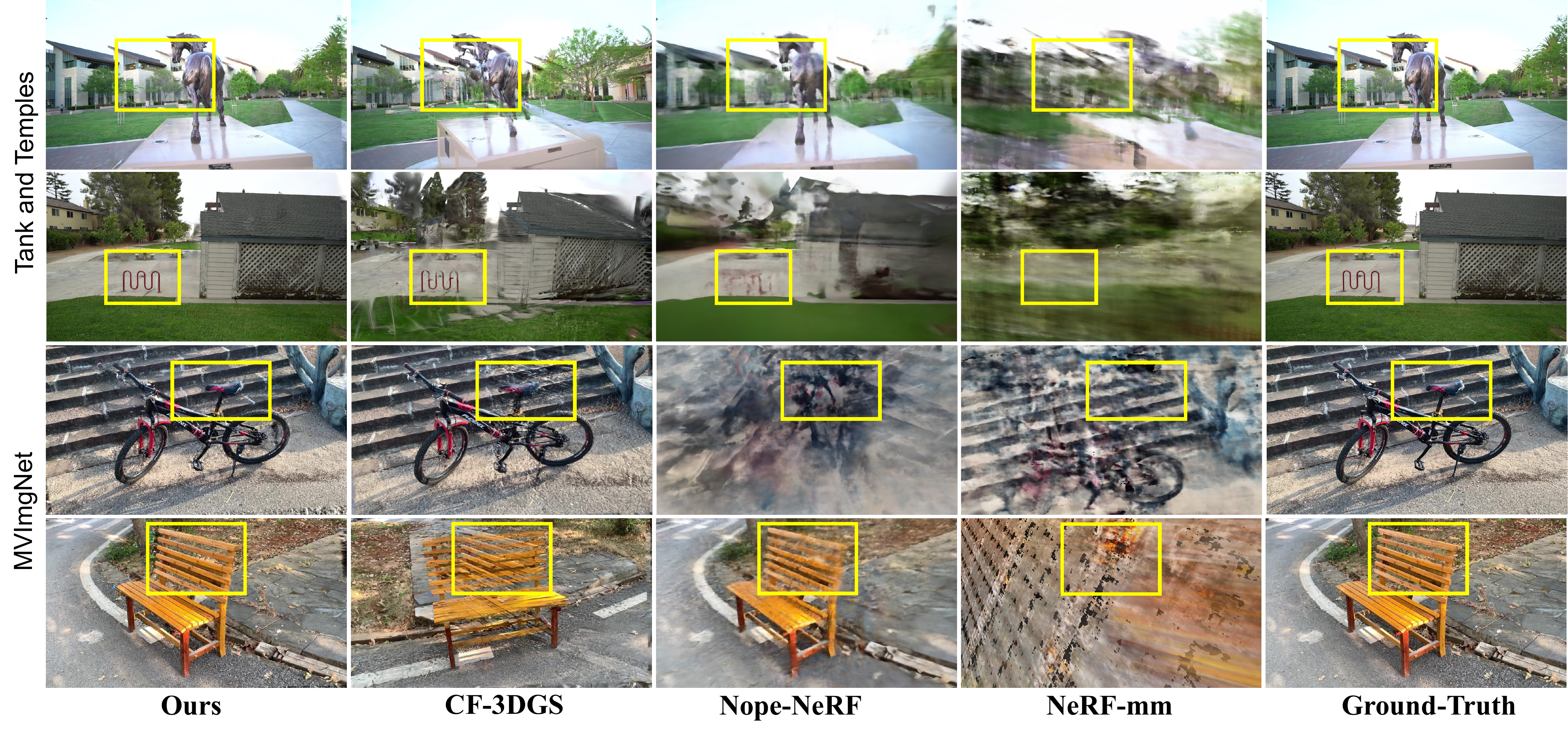}
\caption{\textbf{Visual Comparisons} between InstantSplat and various pose-free baseline methods. InstantSplat achieves faithful 3D reconstruction and renders novel views with as few as three training images on the Tanks and Temples datasets, and the MVImgNet datasets.}
\label{fig:main_exp1}
\end{center}
\end{figure*}

%% file: sec/4_exp.tex
\section{Experiments}

\subsection{Experimental Setup}

\begin{table*}
\centering
\caption{\textbf{Quantitative Evaluations on MVImgNet Datasets.} Our method renders significantly clearer details (by LPIPS) compared to other pose-free methods, devoid of artifacts typically associated with noisy pose estimation (e.g., CF-3DGS~\cite{fu2023colmap}, NeRFmm~\cite{wang2021nerf}) and NoPe-NeRF~\cite{bian2023nope}. 
COLMAP on full views as the ground-truth reference. Our method produces more accurate camera pose estimation in the Absolute Trajectory Error (ATE) than previous COLMAP-free methods, quantified using the ground truth scale.
ADC is denoted as Adaptive Density Control~\cite{kerbl20233d}.
\href{https://pypi.org/project/cupy/}{\textbf{Cupy}} library is used to accelerate the initialization.
}\label{tab:mvimgnet_mipnerf360_eval}
\resizebox{1.99\columnwidth}{!}{
\begin{tabular}{l|ccc|ccc|ccc|ccc}
 &\multicolumn{3}{c|}{Training Time}   & \multicolumn{3}{c|}{SSIM}   & \multicolumn{3}{c|}{LPIPS}  & \multicolumn{3}{c}{ATE}            \\
   & 3-view   & 6-view  & 12-view    & 3-view   & 6-view  & 12-view   & 3-view   & 6-view  & 12-view     & 3-view   & 6-view  & 12-view             
\\ \hline
NoPe-NeRF~\cite{bian2023nope}  
& 37min42s & 53min & 95min & 0.4326 & 0.4329 & 0.4686  & 0.6674  & 0.6614 & 0.6257 & 0.2780 & 0.1740 & 0.1493
\\
CF-3DGS~\cite{fu2023colmap}
  & 3min47s & 7min29s & 10min36s & 0.3414 & 0.3544 & 0.3655 & 0.5523 & 0.4326 & 0.4492 & 0.1593 & 0.1981 & 0.1243
\\\hline
NeRF-mm~\cite{wang2021nerf}                  
& 8min35s & 16min30s & 33min14s & 0.3752 & 0.3685 & 0.3718 & 0.7001 & 0.6252 & 0.6020 & 0.2721 & 0.2376 & 0.1529
\\\hline
\textbf{Our-S}                                                     
& \textbf{10.4s} & \textbf{15.4s} & \textbf{34.1s} &  0.5489 & 0.6835 & 0.7050 & 0.3941 & 0.2980 & 0.3033 & 0.0184 & 0.0259 & 0.0165
\\
\textbf{Our-XL}              
& 35.7s & 48.1s & 76.7s & \textbf{0.5628} & \textbf{0.6933} & \textbf{0.7321} & \textbf{0.3688} & \textbf{0.2611} & \textbf{0.2421} & \textbf{0.0184} & \textbf{0.0259} & \textbf{0.0164}
\\ 
\end{tabular}}
\end{table*}

\paragraph{\textbf{Datasets.}}
Following methodologies from previous pose-free research~\cite{fu2023colmap,bian2023nope}, we evaluated our InstantSplat on eight scenes from the \textbf{Tanks and Temples} dataset \cite{knapitsch2017tanks} with view counts ranging from 3 to 12. Additionally, we tested on seven outdoor scenes from the \textbf{MVImgNet} dataset~\cite{yu2023mvimgnet}, covering diverse scene types including Car, Suv, Bicycle, Chair, Ladder, Bench, and Table. Our experiments also extend to in-the-wild data, incorporating frames from the \textbf{Sora} video~\href{https://openai.com/index/sora/}{(link)}, and navigation stereo camera data from the \textbf{Perseverance Rover}~\href{https://mars.nasa.gov/mars2020/multimedia/raw-images/}{(NASA)}, along with three randomly sampled training views from the DL3DV-10K dataset~\cite{ling2023dl3dv}.

\paragraph{\textbf{Train/Test Datasets Split.}}
For both training and evaluation, 12 images from each dataset were sampled, spanning the entire set. The test set includes 12 images uniformly chosen to exclude the first and last frames, while the training set comprises an equal number of images selected from the remainder, ensuring sparse-view training setting. This uniform sampling approach was applied consistently across both MVImgNet and Tanks and Temples datasets.

\vspace{-2mm}
\paragraph{\textbf{Metrics.}} We evaluate our approach on benchmark datasets~\cite{knapitsch2017tanks,yu2023mvimgnet} across three tasks: novel view synthesis, camera pose estimation, and depth estimation. For novel view synthesis, we use Peak Signal-to-Noise Ratio (PSNR), Structural Similarity Index Measure (SSIM)~\cite{wang2004image}, and Learned Perceptual Image Patch Similarity (LPIPS)~\cite{zhang2018unreasonable}. For camera pose estimation, we report the Absolute Trajectory Error (ATE) as defined in~\cite{bian2023nope}, using COLMAP poses from all dense views as ground-truth references. For depth estimation, we evaluate using the Absolute Relative Error (Rel) and the Inlier Ratio ($\tau$) with a threshold of 1.03, following DUSt3R~\cite{wang2023dust3r}.
\vspace{-2mm}
\paragraph{\textbf{Baselines.}} Our comparisons on pose-free methods include Nope-NeRF~\cite{bian2023nope} and CF-3DGS~\cite{fu2023colmap}, both supported by monocular depth maps and ground-truth camera intrinsics. We also consider NeRFmm~\cite{wang2021nerf}, which involves joint optimization of NeRF and all camera parameters. Additionally, we compare our method with 3D-GS~\cite{kerbl20233d} and FSGS~\cite{zhu2023fsgs}, which utilize COLMAP for pre-computing the camera parameters.

\vspace{-2mm}
\paragraph{\textbf{Implementation Details.}} Our implementation is built using the PyTorch framework. Optimization iterations are set to 200 for \textbf{Ours-S} and 1,000 for \textbf{Ours-XL}, striking a balance between quality and training efficiency. During evaluation, 500 iterations are used for test view optimization. The parameter $\beta$ is fixed at 100. For multi-view stereo (MVS) depth map prediction, MASt3R is configured with a resolution of 512 and the \href{https://pypi.org/project/cupy/}{\textbf{cupy}} library is used to accelerate the initialization. All experiments are conducted on a single Nvidia A100 GPU to ensure fair comparisons.

\begin{table}
\centering
\caption{\textbf{Ablation Study} for validating design choices. We opt for the \textbf{XL} setting, where 1,000 iterations are adopted to ensure reconstruction quality.
Results are averaged on all scenes of Tank and Temple datasets while \href{https://pypi.org/project/cupy/}{\textbf{Cupy}} library is adopted.}
\label{tab:ablation_quant}
\resizebox{0.98\columnwidth}{!}{
\begin{tabular}{lcc|cc|c} \toprule
Model                                   & Train$\downarrow$ & FPS$\uparrow$ & SSIM$\uparrow$  & LPIPS$\downarrow$ & ATE$\downarrow$       \\ \midrule
Baseline by Stitching                   & 51s             & 150           & 0.8014& 0.1654& 0.0142\\ \hline
+ Focal Averaging                & 51s             & 150           & 0.8406& 0.1288& 0.0115\\
+ Co-visibility Geometry Init.   & 45s             & 181           & 0.8411& 0.1313& 0.0115\\
+ Confidence-aware Optimizer            & 45s              & 181           & 0.8553& 0.1277& 0.0115\\
+ Gaussian Bundle Adjustment           & \textbf{45s}     & \textbf{181}  & \textbf{0.8822}& \textbf{0.1050}& \textbf{0.0102}\\ \hline

\end{tabular}}
\vspace{-3mm}
\end{table}

\subsection{Experimental Results}

\paragraph{\textbf{Quantitative and Qualitative Results on Novel-view Synthesis}}
We evaluate novel-view synthesis and pose estimation on the Tanks and Temples dataset, with results summarized in Tab.\ref{tab:tt_eval} and Fig.\ref{fig:main_exp1}.

Nope-NeRF~\cite{bian2023nope}, which leverages Multilayer Perceptrons (MLPs), delivers promising pose estimation accuracy with dense video sequences. However, it often generates overly blurred images (see the third column in Fig.~\ref{fig:main_exp1}), primarily due to the heavy constraints imposed by its geometric field. Additionally, it suffers from prolonged training times for a single scene and inference delays ($\sim$3 seconds per frame for rendering one image).
CF-3DGS~\cite{fu2023colmap}, which employs Gaussian Splatting with local and global optimization stages along with adaptive density control and opacity reset policies, encounters artifacts when rendering from novel viewpoints. These artifacts result from its complex optimization pipeline and erroneous pose estimations, as shown in the second column of Fig.~\ref{fig:main_exp1}. Moreover, both Nope-NeRF and CF-3DGS assume accurate focal lengths, limiting their robustness in scenarios with focal length uncertainties.
NeRFmm, designed for simultaneous optimization of camera parameters and the radiance field, often struggles with suboptimal results due to the inherent challenges of naive joint optimization.

Pose metrics, as reflected in Tab.~\ref{tab:tt_eval}, reveal significant artifacts caused by sparse observations and inaccurately estimated poses. These limitations are particularly pronounced in CF-3DGS and Nope-NeRF, which depend on dense video sequences, akin to SLAM, and encounter difficulties when these dense frames are downsampled to sparse multi-view images.
In contrast, our method initializes with the MVS scene structure and employs a gradient-based joint optimization framework, providing enhanced robustness and superior performance.

We demonstrate the geometric accuracy of InstantSplat by integrating it with 2D Gaussian Splatting, visualizing rasterized depth and normal maps. As shown in Fig.~\ref{fig:main_exp_geo} and Tab.~\ref{tab:is_2D-GS_nvs_depth}, under the 3-view setting, InstantSplat significantly improves geometry quality, benefiting from dense initialization and the joint optimization framework. Additionally, we validate InstantSplat within the multi-resolution Mip-Splatting framework. Tab.~\ref{tab:mip} shows evaluations on the Tanks and Temples dataset using the original resolution for testing while varying the training resolution. InstantSplat consistently enhances sparse-view rendering metrics.

Further experiments on the MVImgNet dataset with 3, 6, and 12 training images are presented in Tab.~\ref{tab:mvimgnet_mipnerf360_eval}. This section demonstrates that InstantSplat consistently outperforms all baselines across all evaluated datasets and metrics. Please refer to our supplementary video for additional visualizations, including in-the-wild test cases such as the Sora video, DL3DV-10K dataset, and stereo image pairs from the Mars rover.

\subsection{Ablations and Analysis}
We conducted ablation studies to validate our design choices, transitioning from the use of non-differentiable Structure-from-Motion (SfM) complemented by Gaussian Splatting with adaptive density control, to adopting Multi-View Stereo (MVS) with Co-visible Global Geometry Initialization, Confidence-aware Optimizer and Gaussian Bundle Adjustment for efficient and robust sparse-view 3D modeling. Experiments were conducted using 12 training views on all scenes in the Tanks and Temples dataset, with COLMAP poses derived from full dense views serving as ground truth, unless otherwise specified.
\begin{itemize}
    \item Question 1: How does the quality of multi-view stereo predictions impact performance?
    \item Question 2: What is the impact of focal averaging on the results?
    \item Question 3: Does co-visible geometric initialization effectively reduce redundancy and preserve reconstruction accuracy?
    \item Question 4: Is Gaussian Bundle Adjustment essential for accurate rendering?
    \item Question 5: Can InstantSplat achieve rendering quality comparable to methods using dense-view images, but with sparse-view inputs?
\end{itemize}

\noindent For \textbf{Q1}, the evaluation of camera pose accuracy from MASt3R reveals significant discrepancies in rendering quality and pose accuracy, as shown in Tab.~\ref{tab:ablation_quant}'s first and last rows.

\noindent For \textbf{Q2}, aggregating focal lengths from all images stabilizes the 3D representation optimization, evidenced by results in the second row.

\noindent For \textbf{Q3}, co-visible geometry initialization markedly improves FPS by reducing redundancy. Confidence-based view ranking further enhances rendering accuracy, as detailed in the third and fourth rows.

\noindent For \textbf{Q4}, Gaussian Bundle Adjustment (GauBA), through self-supervised optimization, substantially improves visual and pose metrics by aligning the 3D and 2D representations.

\noindent For \textbf{Q5}, InstantSplat using 12 training images approaches the rendering quality of denser-view methods as measured by the Learned Perceptual Image Patch Similarity (LPIPS), indicative of near-human perceptual sharpness. However, it exhibits discrepancies in pose estimation accuracy (ATE), inevitably leading to lower PSNR metrics (Tab.~\ref{tab:full_view}).

\begin{table}[t]
\centering
\caption{Performance comparison between InstantSplat (ours) with 12 training views, CF-3DGS and 3D-GS with dense (100-400) training views on the Tanks and Temples dataset. We report the LPIPS, reconstruction time, and pose accuracy.}\label{tab:full_view}
\resizebox{0.98\columnwidth}{!}{
\begin{tabular}{@{}l|cccc}
 & Views~$\downarrow$ & Time~$\downarrow$ & LPIPS~$\downarrow$  & ATE $\downarrow$\\
\hline
CF-3DGS~\cite{fu2023colmap} & 100-400 & $\sim$30min & 0.09  & 0.004  \\
COLMAP+3DGS~\cite{kerbl20233d} & 100-400 & $\sim$30min  & 0.10  & - \\
InstantSplat (Ours) & 12 & $\sim$45s  & 0.10  & 0.010   \\
\end{tabular}%
}
\label{tab:dtu_perf}
\end{table}

\begin{table}
\centering
\caption{\textbf{InstantSplat with 2D-GS}. InstantSplat demonstrates strong compatibility with 2D-GS, notably improving sparse-view rendering quality and depth accuracy while also accelerating training by eliminating the need for an additional SfM step. \href{https://pypi.org/project/cupy/}{\textbf{Cupy}} library is used to accelerate the initialization.}\label{tab:is_2D-GS_nvs_depth}
\resizebox{0.98\columnwidth}{!}{
\begin{tabular}{c|c|c|ccc|cc}
                               & \multirow{2}{*}{\begin{tabular}[c]{@{}c@{}}View~\\Num.\end{tabular}} & \multirow{2}{*}{\begin{tabular}[c]{@{}c@{}}Train~\\Time\end{tabular}} &  \multicolumn{3}{c|}{Novel View Synthesis} & \multicolumn{2}{c}{Depth Estimation} \\ \cline{4-8}
                               &          &                                     & PSNR$\uparrow$      & SSIM$\uparrow$     & LPIPS$\downarrow$    & rel$\downarrow$      & $\tau$$\uparrow$     \\ \hline
\multirow{3}{*}{\begin{tabular}[c]{@{}c@{}}2D-GS~\\+InstantSplat\end{tabular}} & 3      &       45s                                & 21.90   & 0.7447   & 0.1920   & 28.7557 & 26.1918   \\
                               & 6        &        58s                              & 24.99& 0.8281& 0.1450& 29.9461& 25.7376   \\
                               & 12       &        98s                              & 26.07   & 0.8557  & 0.1407   & 29.4234   & 25.3569   \\ \hline
\multirow{3}{*}{2D-GS}          & 3        &        10min3s                              & 15.09   & 0.5495   & 0.3719   & 59.8629   & 5.2962   \\
                               & 6        &        11min22s                              & 19.88   & 0.7251   & 0.2388   & 45.4608   & 17.8145   \\
                               & 12       &        11min54s                              & 23.70   & 0.8184   & 0.1723   & 40.1948   & 20.1998   \\ \hline
\end{tabular}}
\end{table}

\begin{table}
\centering
\caption{\textbf{InstantSplat with Mip-Splatting}. Both methods were trained in sparse view settings, with variations in training and evaluation resolutions consistent with Mip-Splatting protocols. InstantSplat demonstrates compatibility with Mip-Splatting, significantly enhancing sparse-view rendering quality.}\label{tab:mip}
\resizebox{0.98\columnwidth}{!}{
\begin{tabular}{l|c|cc|cc}
                               & \multirow{2}{*}{\begin{tabular}[c]{@{}c@{}}Training~\\Resolution\end{tabular}} & \multicolumn{2}{c|}{3-view} & \multicolumn{2}{c}{12-view}  \\ \cline{3-6}
                               &                                                                                & SSIM   & LPIPS              & SSIM   & LPIPS               \\ \hline
\multirow{3}{*}{Mip-Splatting + Ours-XL}          & 1                                                                              & 0.7647 & 0.1618             & 0.8415 & 0.1305              \\
                               & $\sfrac{1}{2}$ \text{Res.}                                                                           & 0.7456 & 0.2343             & 0.7945 & 0.2027              \\
                               & $\sfrac{1}{4}$ \text{Res.}                                                                            & 0.6630 & 0.3774             & 0.7204 & 0.3348              \\ \hline
\multirow{3}{*}{Mip-Splatting} & 1                                                                              & 0.5422 & 0.3780             & 0.7947 & 0.1901              \\
                               &$\sfrac{1}{2}$ \text{Res.}                                                                            & 0.5777 & 0.3939             & 0.7703 & 0.2418              \\
                               & $\sfrac{1}{4}$ \text{Res.}                                                                           & 0.5477 & 0.4840             & 0.6664 & 0.3770              \\ \hline
\end{tabular}}
\end{table}

%% file: sec/5_conclusion.tex
\section{Conclusion}
InstantSplat is a system tailored to reconstruct large-scale scenes from sparse-view, unposed images. Operating within a self-supervised framework, InstantSplat optimizes camera poses and 3D representations using photometric signals and avoids the time-consuming Adaptive Density Control. The system leverages multi-view stereo priors for surface point initialization and employs co-visible cross-view checking to minimize model redundancy. Compared with previous best-performing pose-free methods~\cite{fu2023colmap,bian2023nope}, InstantSplat dramatically reduces the required number of views from hundreds to just a few and demonstrates robustness across in-the-wild datasets, as well as adaptability to various point-based representations.

However, InstantSplat faces limitations when applied to scenes with more than hundreds of images due to its reliance on MVS for globally aligned point clouds. Addressing this through progressive alignment techniques will be a focus of future research.

%% file: sec/X_suppl.tex
\renewcommand{\thepage}{A\arabic{page}}
\renewcommand{\thesection}{A\arabic{section}}
\renewcommand{\thetable}{A\arabic{table}}   
\renewcommand{\thefigure}{A\arabic{figure}}

\section{Details of Global Geometry Initialization} \label{sec:more_details}

We provide additional details on the scaling process for pairwise-to-global alignment as discussed in Co-visible Global Geometry Initialization (Section 3.3 of the main draft).

Specifically, the geometric prior model, MASt3R, originally uses image pairs as input, requiring post-processing to align scales when more than two views are captured from the scene. This necessity arises because the predicted point maps are generated at varying scales. Such scale misalignment across independently computed relative poses introduces variance, leading to inaccuracies in camera pose estimation.

To integrate these pairwise pointmaps and camera information into a globally aligned geometry, we follow MASt3R~\cite{leroy2024grounding} to construct a complete connectivity graph $\mathcal{G}(\mathcal{V},\mathcal{E})$ of all the N input views, where the vertices $\mathcal{V}$ and each edge $e=(n,m) \in \mathcal{E}$ indicate a shared visual content between images $I_n$ and $I_m$. To convert the initially predicted point map $\{ (\boldsymbol{P}_i \in \mathbb{R}^{H \times W \times 3}) \}_{i=1}^{N}$ to be a globally aligned one $\{ (\boldsymbol{\tilde{P}}_i \in \mathbb{R}^{H \times W \times 3}) \}_{i=1}^{N}$, we update the point maps, transformation matrix, and a scale factor:
For the complete graph, any image pair $e=(n,m) \in \mathcal{E}$, the point maps $\boldsymbol{P}_{n,n}$,$\boldsymbol{P}_{m,n}$ and confidence maps $\boldsymbol{O}_{n,n}$,$\boldsymbol{O}_{m,n}$. For clarity, let us define $\boldsymbol{P}_{n,e} := \boldsymbol{P}_{n,n}$, and $\boldsymbol{P}_{m,e} := \boldsymbol{P}_{m,n}$, $\boldsymbol{O}_{n,e} := \boldsymbol{O}_{n,n}$, and $\boldsymbol{O}_{m,e} := \boldsymbol{O}_{m,n}$.
The optimization for the transformation matrix of edge $\boldsymbol{T}_e$, scaling factor $\sigma_e$ and point map $\tilde{\boldsymbol{P}}$ are given by:

\begin{equation}
    \tilde{\boldsymbol{P}}^{*} = \underset{\tilde{\boldsymbol{P}},\boldsymbol{T},\sigma}{\arg\min} \sum_{e \in \mathcal{E}} \sum_{v \in e} \sum_{i=1}^{HW}
    \boldsymbol{O}_{v,e}^i \left\Vert \tilde{\boldsymbol{P}}_{v}^{i} - \sigma_e \boldsymbol{T}_e \boldsymbol{P}_{v,e}^{i} \right\Vert.
\label{eq:global_align}
\end{equation}

Here, we slightly abuse notation and use $v\in e$ for $v \in \{n,m\}$ if $e=(n,m)$.
The idea is that, for a given pair $e$, the same rigid transformation $\boldsymbol{T}_e$ should align both pointmaps $\boldsymbol{P}_{n,e}$ and $\boldsymbol{P}_{m,e}$ with the world-coordinate pointmaps $\tilde{\boldsymbol{P}}_{n,e}$ and $\tilde{\boldsymbol{P}}_{m,e}$,
since $\boldsymbol{P}_{n,e}$ and $\boldsymbol{P}_{m,e}$ are by definition both expressed in the same coordinate frame.
To avoid the trivial optimum where $\sigma_e=0,\, \forall e\in\mathcal{E}$, we enforce $\prod_e \sigma_e=1$.
Having aligned the point clouds, we commence the integration process by initializing 3D Gaussians~\cite{kerbl20233d}, treating each point as a primitive.
This post-processing step yields a globally aligned 3D Gaussians within only seconds, where inferring per-view point and confidence maps can be achieved real-time on a modern GPU.

\subsection{Additional Implementation Details.} We implement InstantSplat-2D-GS using the original 2D-GS CUDA kernel. All core components of our system are integrated into 2D-GS, including Focal Averaging, co-visible Global Geometry Initialization, Confidence-aware Optimization, and Gaussian Bundle Adjustment. The optimization iterations are set to 1,500 for Gaussian training and 500 for aligning test view camera poses. Regularization for depth distortion and normal consistency begins at iterations 500 and 700, respectively.

\subsection{Evaluation on 2D-GS}
In Table~\ref{tab:is_2D-GS_nvs_depth}, we present the performance of InstantSplat-2D-GS on the tasks of novel view synthesis and multi-view stereo depth estimation using the Tanks and Temples dataset~\cite{knapitsch2017tanks}. Following the train/test dataset split scheme described in the main draft, we sample 12 images for training and 12 for evaluation, covering the entire dataset. The test set contains 12 uniformly selected images (excluding first/last frames), with an equal number of training images sampled from remaining frames for sparse-view training.
For novel view synthesis, we evaluate performance using PSNR, SSIM, and LPIPS. For depth estimation, we use the Absolute Relative Error (Rel) and Inlier Ratio ($\tau$) with a threshold of 1.03 to assess each scene, following the approach of DUSt3R~\cite{wang2023dust3r}. We align the scene scale between the predictions and the ground truth, as our system does not depend on camera parameters for prediction. We normalize the predicted depth maps using the median of the predicted depths and similarly normalize the ground truth depths, which follows the procedures established in previous studies~\cite{schroppel2022benchmark, wang2023dust3r} to ensure alignment between the predicted and ground truth depth maps. 
We use COLMAP~\cite{schonberger2016structure} to perform dense stereo reconstruction on the entire video sequence, generating a high-quality multi-view stereo point cloud. The 3D dense points are then projected onto 2D image planes to create pseudo ground-truth depth maps for evaluation.

\section{More Experimental Results} \label{sec:more_exp_results}
\subsection{Additional Qualitative Results on the Benchmarks}
\begin{figure*}[t!]
\begin{center}
\centering
\includegraphics[width=0.99\linewidth]{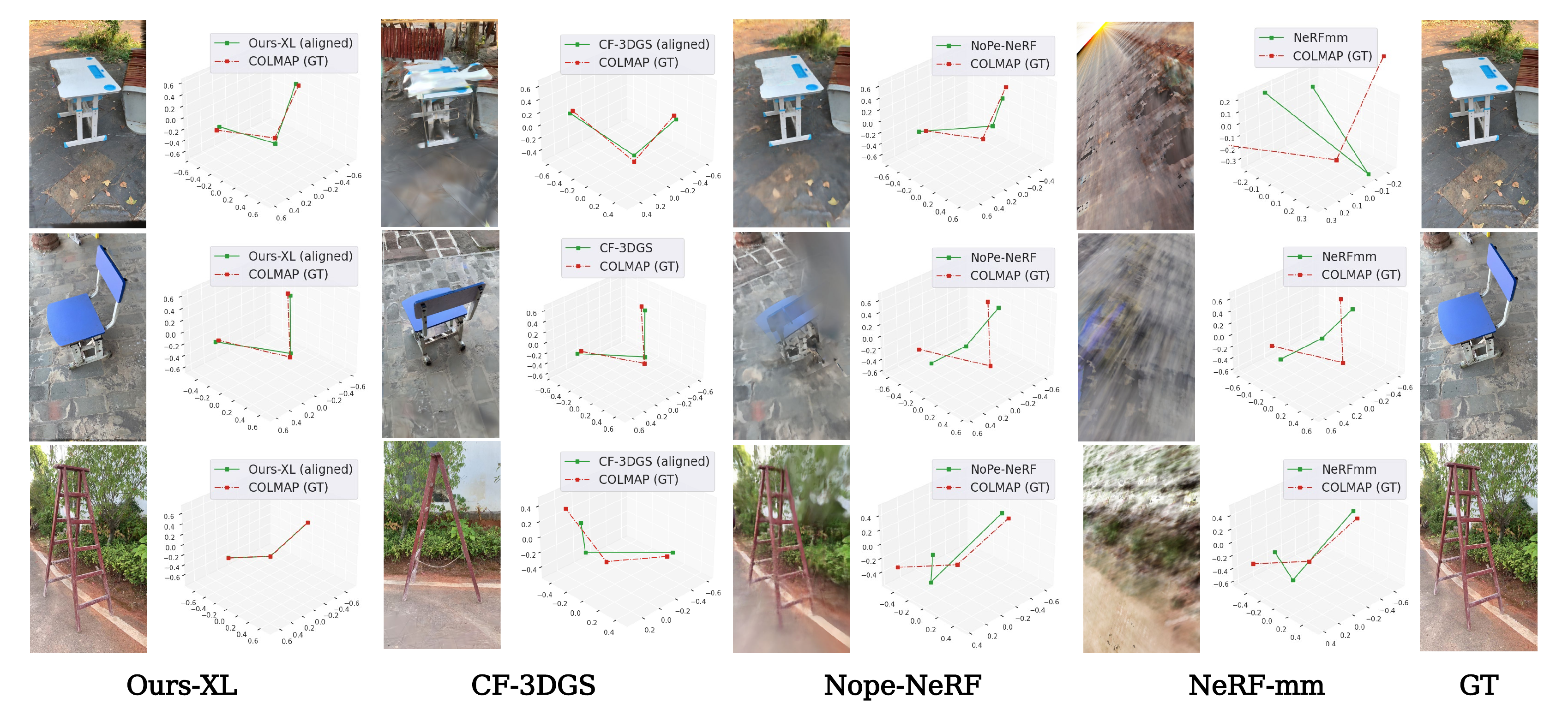}
\caption{\textbf{Additional Visual Comparisons on the MVimgNet Dataset}. We present the rendering results from NeRFmm, Nope-NeRF, CF-3DGS, and Ours-XL.}
\label{fig:supp_fig3}
\end{center}
\end{figure*}

We provide more visualized results and pose estimations on realistic outdoor scenes from the Tanks and Temples dataset~\cite{knapitsch2017tanks} and MVImgNet~\cite{yu2023mvimgnet}. As shown in Fig.\ref{fig:supp_fig2} and Fig.\ref{fig:supp_fig3}, InstantSplat consistently produces fewer artifacts than Gaussian-based methods (e.g., CF-3DGS~\cite{fu2023colmap}) and exhibits sharper details than NeRF-based methods (e.g., Nope-NeRF~\cite{bian2023nope} and NeRFmm~\cite{wang2021nerf}).

\subsection{Video Demonstration} \label{sec:webpage}

\begin{figure*}[h]
\begin{center}
\centering
\includegraphics[width=0.99\linewidth]{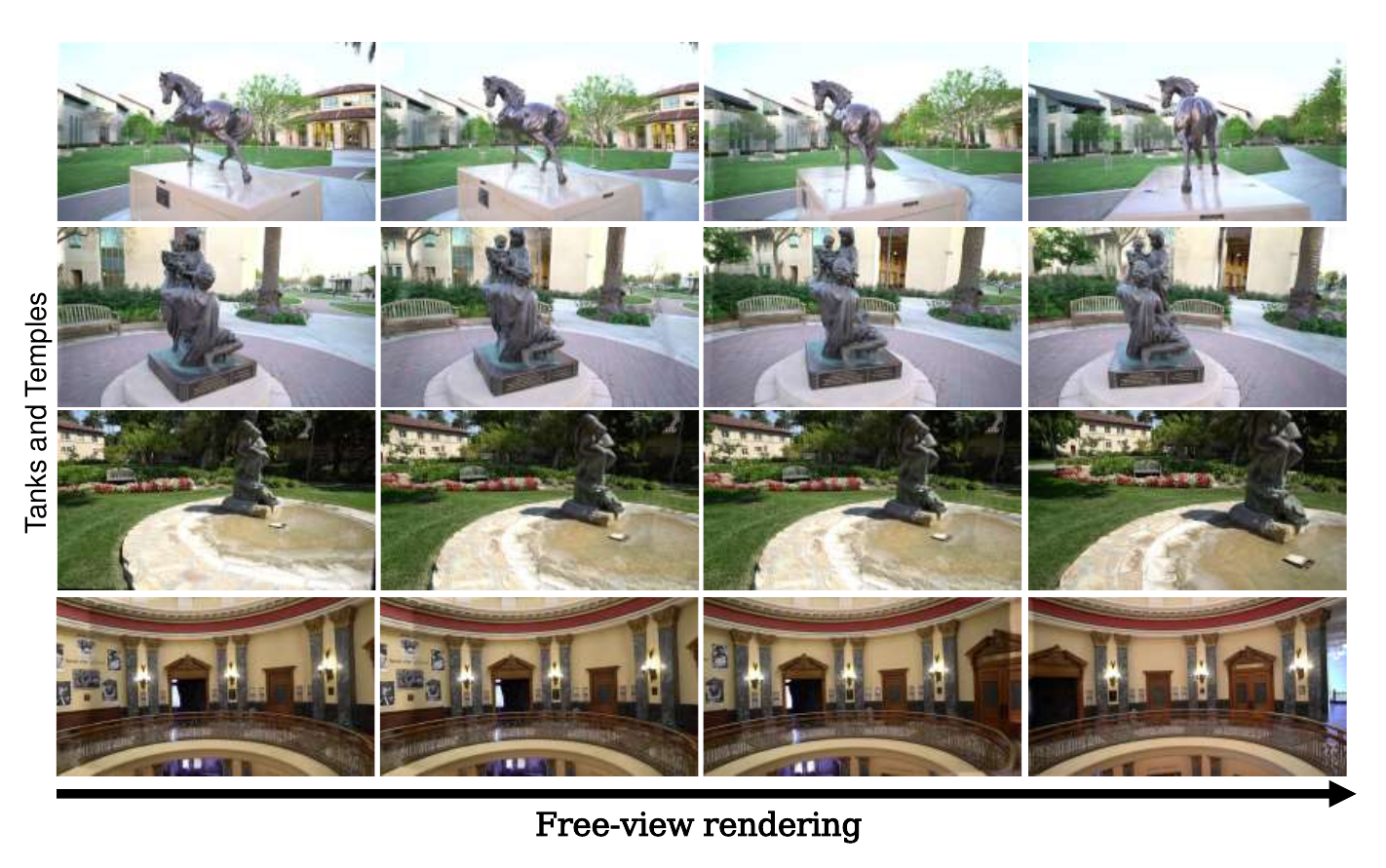}
\caption{\textbf{Free-view Rendering on the Tanks and Temples Dataset.}}
\label{fig:supp_fig1}
\end{center}
\end{figure*}

We provide free-viewpoint rendering on various in-the-wild datasets, including the~\href{https://openai.com/index/sora/}{\textbf{Sora}} video, by uniformly sampling three frames from the entire video sequence. Additionally, we use stereo camera data from the \textbf{Perseverance Rover}, available on the~\href{https://mars.nasa.gov/mars2020/multimedia/raw-images/}{NASA website}, and randomly sample three training views from the DL3DV-10K dataset~\cite{ling2023dl3dv}. As shown in Fig.~\ref{fig:supp_fig1} and Fig.~\ref{fig:supp_fig2}, InstantSplat achieves accurate 3D reconstruction and high-quality novel view synthesis with as few as three training images, highlighting the robustness and effectiveness of our framework. For additional visualizations, please refer to our \underline{instantsplat-webpage.zip} with opening the ``index.html''.

\begin{figure*}[t!]
\begin{center}
\centering
\includegraphics[width=0.92\linewidth]{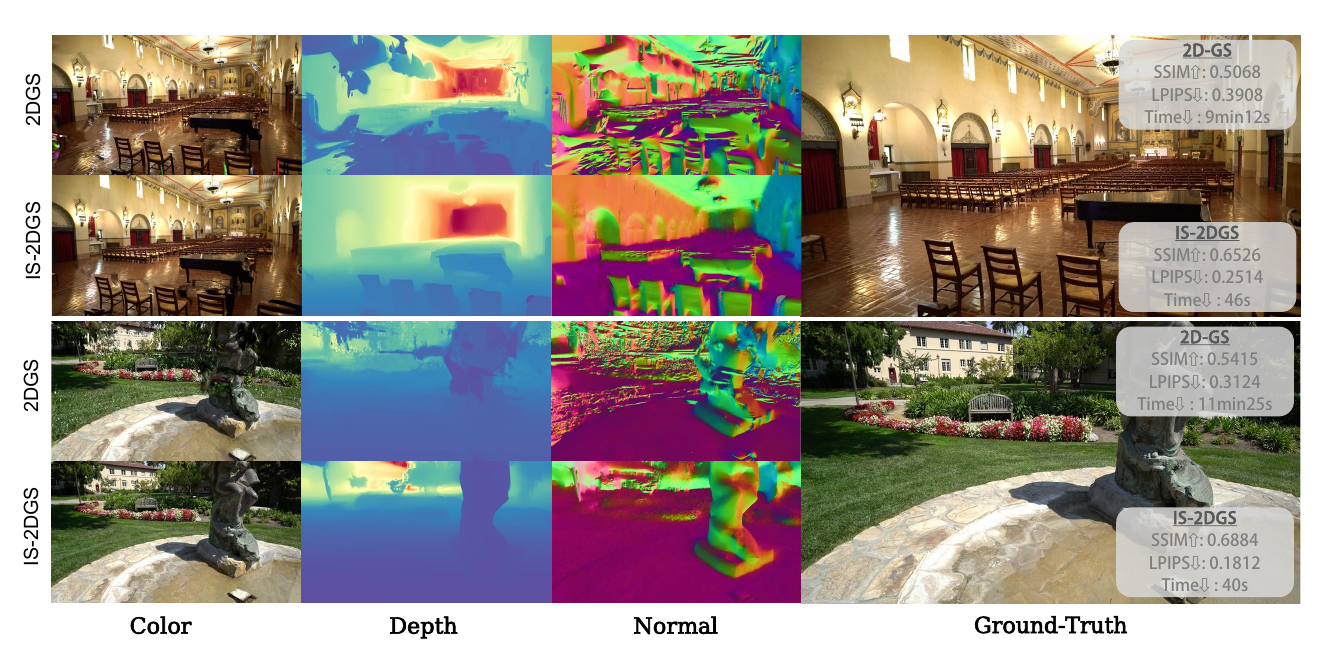}
\caption{\textbf{Additional Visual Comparisons between 2D-GS and InstantSplat-2D-GS.} We present the rendering results of RGB images, depth maps, and normal maps from the Tanks and Temples dataset. The corresponding SSIM, LPIPS, and training time metrics are displayed on the right. Here, IS is used as an abbreviation for InstantSplat.}
\label{fig:depth_normal}
\end{center}
\end{figure*}

\begin{figure*}[h]
\begin{center}
\centering
\includegraphics[width=0.99\linewidth]{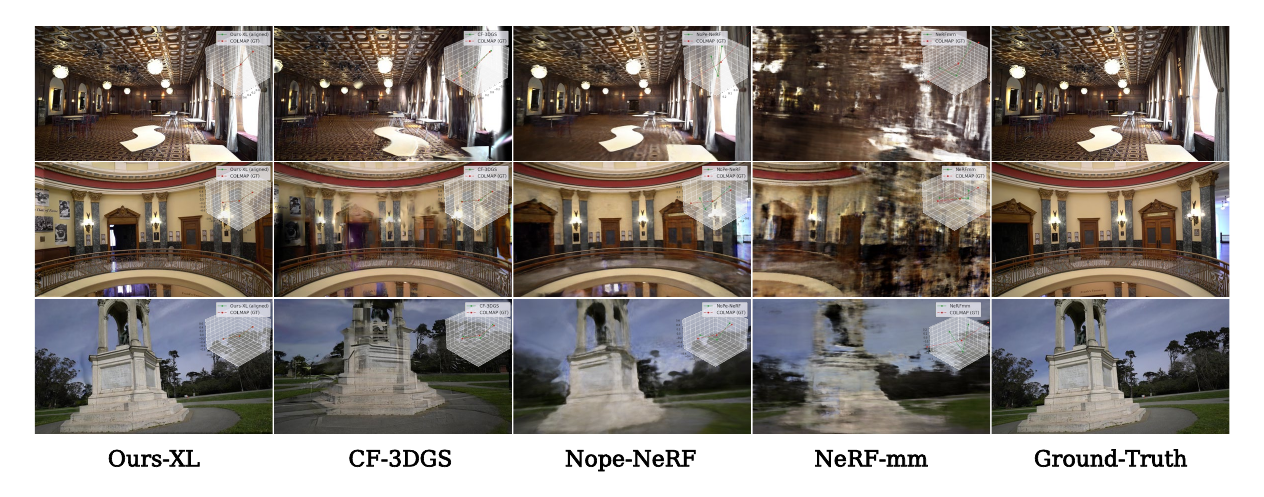}
\caption{\textbf{Additional Visual Comparisons on the Tanks and Temples Dataset}. We present the rendering results from NeRFmm, Nope-NeRF, CF-3DGS, and Ours-XL.}
\label{fig:supp_fig2}
\end{center}
\end{figure*}

\begin{figure*}[h]
\begin{center}
\centering
\includegraphics[width=0.99\linewidth]{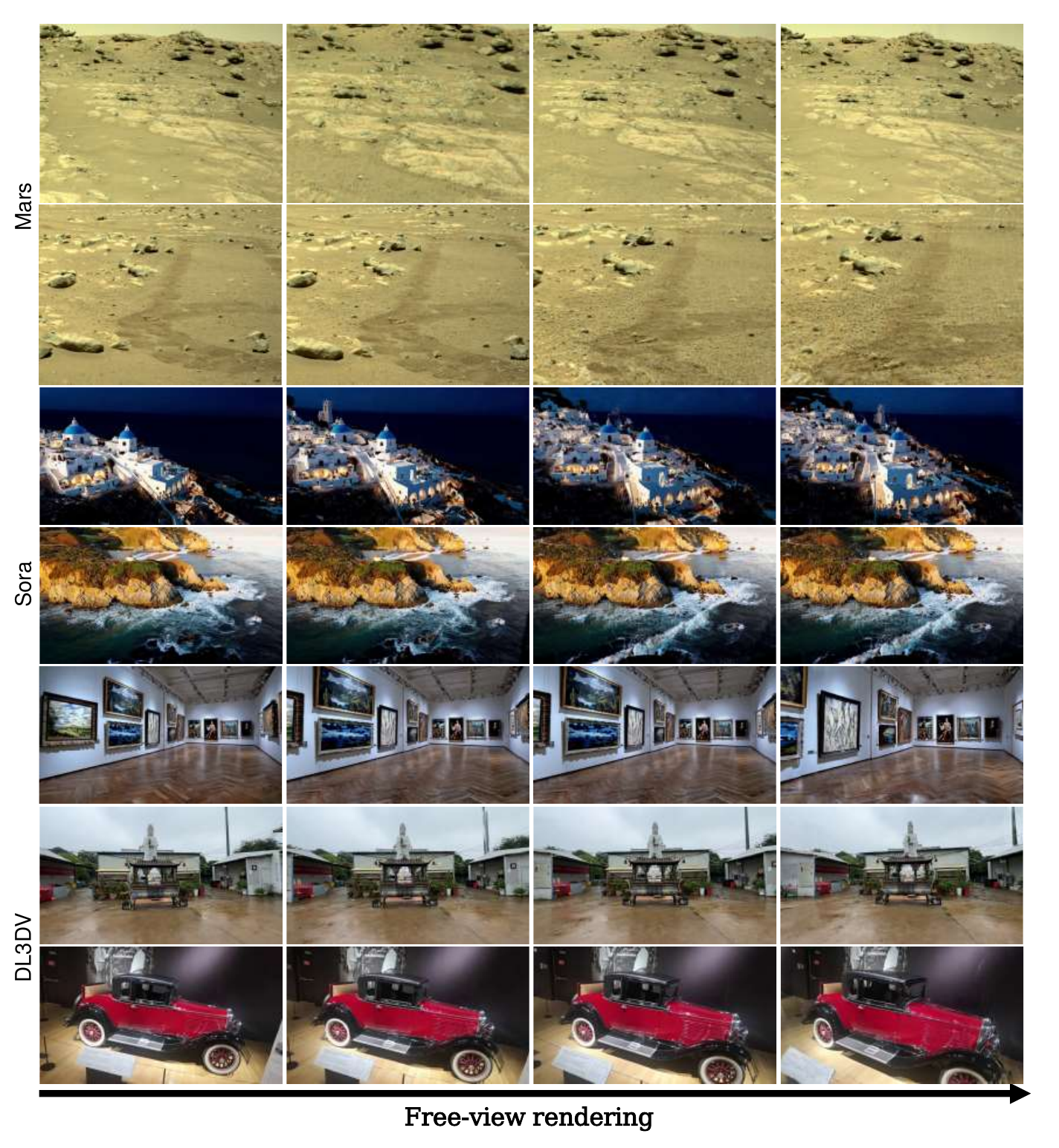}
\caption{\textbf{Free-view Rendering on in-the-wild data.} We present the rendering results of three different in-the-wild data: Mars Rover Navigation, Sora Video and DL3DV-10K dataset}
\label{fig:supp_fig2}
\end{center}
\end{figure*}

\clearpage